\documentclass{article}

\usepackage[final]{nips_2017}

\usepackage[utf8]{inputenc} % allow utf-8 input
\usepackage[T1]{fontenc}    % use 8-bit T1 fonts
\usepackage{hyperref}       % hyperlinks
\usepackage{url}            % simple URL typesetting
\usepackage{booktabs}       % professional-quality tables
\usepackage{amsfonts}       % blackboard math symbols
\usepackage{nicefrac}       % compact symbols for 1/2, etc.
\usepackage{microtype}      % microtypography
\usepackage{multirow}

\usepackage{algorithm,algorithmic}
\usepackage{amsmath}
\usepackage{graphicx}
\usepackage{booktabs}
\usepackage{subcaption}

\usepackage[colorinlistoftodos]{todonotes}
\usepackage{color}

\newcommand{\real}{{\mathbb R}}
\usepackage{wrapfig}

\title{Sequential Dynamic Decision Making with Deep Neural Nets on a Test-Time Budget}

\author{
	Henghui Zhu \thanks{Systems Engineering, Boston University}\\
	\texttt{henghuiz@bu.edu}\\
 	\And 		
	Feng Nan \footnotemark[1]\\
	\texttt{fnan@bu.edu} \\
	\And	
	Ioannis Paschalidis \thanks{Electrical Engineering, Boston University}\\
	\texttt{yannisp@bu.edu} \\
	\And
	Venkatesh Saligrama \footnotemark[2]\\
	\texttt{srv@bu.edu} \\
}

\begin{document}
\maketitle

\begin{abstract}
Deep neural network (DNN) based approaches hold significant potential for reinforcement learning (RL) and have already shown remarkable gains over state-of-art methods in a number of applications. The effectiveness of DNN methods can be attributed to leveraging the abundance of supervised data to learn value functions, Q-functions, and policy function approximations without the need for feature engineering. Nevertheless, the deployment of DNN-based predictors with very deep architectures can pose an issue due to computational and other resource constraints at test-time in a number of applications. We propose a novel approach for reducing the average latency by learning a computationally efficient gating function that is capable of recognizing states in a sequential decision process for which policy prescriptions of a shallow network suffices and deeper layers of the DNN have little marginal utility. The overall system is adaptive in that it dynamically switches control actions based on state-estimates in order to reduce average latency without sacrificing terminal performance. We experiment with a number of alternative loss-functions to train gating functions and shallow policies and show that in a number of applications a speed-up of up to almost 5X can be obtained with little loss in performance.     
\end{abstract}

\section{Introduction}

Deep neural networks are being increasingly utilized in a number of sequential decision problems, including video gaming~\cite{mnih2015human}, robotic manipulation~\cite{lillicrap2015continuous,mnih2016asynchronous} and self-driving cars \cite{BojarskiTDFFGJM16}, realizing significant performance gains over more conventional approaches. A key contributor to these remarkable gains can be attributed to the availability of large-scale supervised data in these applications. This has allowed the composition of highly complex approximations for learning $Q$-functions and enabled deep learning approaches to closely approximate complex policies that generalize knowledge and predict suitable actions for different points in the state space. 

On the other hand, the deployment of DNNs in many of these applications faces significant hurdles due to computational and other resource constraints. This cost, often called the \emph{test-time cost}, has increased rapidly for many tasks with ever-growing demands for improved performance in state-of-the-art systems. As a case in point the \textit{Resnet152} \cite{he2016deep} architecture with 152 layers, realizes a substantial 4.4\% accuracy gain in top-5 performance over GoogLeNet \cite{szegedy2015going} on the large-scale ImageNet dataset \cite{russakovsky2015imagenet} but is about 14X slower at test-time. 
The high test-time cost of state-of-the-art DNNs means that they can only be deployed on powerful computers, equipped with massive GPU accelerators. As a result, technology companies spend billions of dollars a year on expensive and power-hungry computer hardware. Moreover, high test-time cost prevents DNNs from being deployed on resource constrained platforms, such as those found on Internet of Things (IoT) devices, smartphones, and wearables. These issues are further exacerbated in sequential decision making applications such as remote navigation and video gaming where there is little room or tolerance for latency and slow response time. 

We propose a novel approach for reducing average latency by learning a computationally efficient gating function to deal with computational constraints. Our gating function, at test-time, identifies states where a shallow neural network (SNN) suffices to predict a good high-value action. If SNN suffices, we use it to predict the action, thereby reducing latency; otherwise a DNN is utilized. Thus, our model is adaptive and dynamically controls which one of the two policies options (SNN or DNN) to use. This general idea is applicable to any sequential decision making process and is based on recognizing situations where actions predicted by a SNN are as good as the DNN. The overall policy function is adaptive and dynamically switches control actions based on state-estimates in order to reduce average latency without sacrificing the overall performance.

We train adaptive policy functions models on annotated training data. In the first step we train a high-complexity DNN following existing architectures \cite{mnih2016asynchronous,mnih2015human,lillicrap2015continuous,van2016deep} that lead to high-quality $Q$-functions and policy functions. Our next step is to jointly train a low-cost gating function together with a shallow neural network (SNN). This step involves identifying input states for which SNN suffices. We formulate an approximation objective that attempts to imitate the high-cost DNN on some suitable partition of the state space. The gating function identifies the partition and the SNN mimics actions of the DNN on the partition. We experiment with different loss functions and approximations ranging from entropy-based policy imitation to alternative adaptive approximation approach with surrogate losses. We experiment on a number of real-world benchmark datasets and demonstrate significant speedups with little loss in performance. 

\section{Background: Deep Reinforcement Learning}
Let $(\mathcal{X},\mathcal{A}, \mathbf{P}, R)$ denote a Markov Decision Process (MDP) with a finite set of states $\mathcal{X}$ and a finite set of actions $\mathcal{A}$.  For any state-action pair $(x, a) \in \mathcal{X}\times \mathcal{A}$, let $P(y |x,a)$ denote the conditional transition probability from state $x$ to state $y \in \mathcal{X}$ after taking action $a$. A policy $\pi(a|x)$ is a map that associates every state with the probability of taking action $a$. The reward function $r: \mathcal{X} \times \mathcal{A} \to \real$ is a function that maps each state-action pair to a real number. The goal of reinforcement learning is to find the policy that maximizes the discounted long-term reward $R = \sum_{k=0}^\infty \gamma^k r(x_{k}, a_{k})$, where $0 < \gamma < 1$ is a discount factor. The {\em value function} $V(x)$ for a state $x$ is the maximum long-term reward obtained starting from $x$. The $Q$-value function $Q(x, a)$ for state-action pair $(x,a)$ is the maximum expected long-term reward obtained starting from $x$ and selecting as first action $a$. The $Q$-value function also satisfies the Bellman equation \cite{bert-dp}, namely,
\begin{equation} \label{Qbellman}
Q(x,a) = r(x,a) + \gamma \sum_{y \in \mathcal{X}} p(y|x, a) \max_u Q(y, u).
\end{equation}
From the solution, say $Q^*(\cdot,\cdot)$, of \eqref{Qbellman} one can also obtain the optimal action at each state $x$ that maximizes $Q^*$. With of loss of generality, we write $\pi(x) = [\pi(a_1|x), \pi(a_2|x), \cdots, \pi(a_n|x)]_{a_i \in \mathcal{A}}$.

Traditional reinforcement learning uses either a lookup table for keeping the $Q$-values, which is memory inefficient and time consuming, or a linear function approximator, which requires appropriate selection of features in order to be effective \cite{bert-dp}. DNNs offer an alternative solution for automatically learning approximations of the value function, or the $Q$-value function, or even the policies directly. Some popular reinforcement learning algorithms can be categorized as actor-learning, critic-learning and actor-critic learning algorithms. Actor-learning iterates on the policy such as the deep policy gradient \cite{lillicrap2015continuous}. Critic-learning iterates on the value function or the $Q$-value function such as DQN \cite{mnih2015human} and Double-DQN \cite{van2016deep}. Actor-critic learning iterates on both the policy and the value function such as Asynchronous Actor-Critic Algorithm (A3C) \cite{mnih2016asynchronous}.

\section{Approach}
We assume a good deep reinforcement learning agent $\pi_{0}$ is either known a priori, or we train to the best possible performance using a very deep architecture. 
\begin{wrapfigure}{h}{0.5\textwidth}
\includegraphics[width=0.5\textwidth]{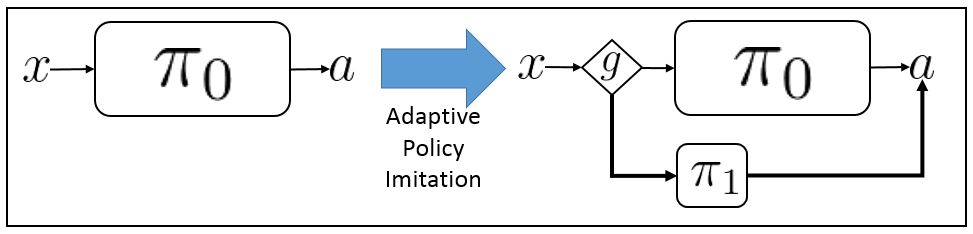}
\caption{A schematic of our approach . $\pi_0$ is a high-accuracy, high-cost policy. We learn low-cost gating $g$ and policy $\pi_1$ to adaptively imitate $\pi_0$.} \label{fig:adaptive_approximation}
\end{wrapfigure}
The evaluation of this good policy $\pi_{0}(x)$ involves many convolutional and fully-connected layers, which can be prohibitively slow in many time-sensitive applications. Let $c(\pi_0)$ denote the computational time required to evaluate an input using $\pi_0$. To meet an average  budget constraint, we propose to learn an alternative weak (low-cost) policy $\pi_{1}$ together with a gating function $g: \mathcal{X} \to \{0, 1\}$ to adaptively approximate $\pi_0$. Given an input $x$, if $g(x)=1$, $\pi_1$ should be employed to obtain the action; if $g(x)=0$, $\pi_0$ should be employed instead. Denote the resulting $(g,\pi_1,\pi_0)$ composite policy as $\pi_{g,1,0}$. The procedure is shown in Figure \ref{fig:adaptive_approximation}. Both $\pi_1$ and $g$ should be fast to evaluate. Let $c(g,\pi_0)$ and $c(g,\pi_1)$ denote the computational time required for evaluating $g+\pi_0$ and $g+\pi_1$, respectively. Our goal is to learn $g$ and $\pi_1$ such that $\pi_{g,1,0}$ has a similar performance to the original good policy $\pi_0$ for the MDP, yet satisfies average cost constraints $\mathbb{E}_{x \sim \mathcal{D}_{\pi_{g,1,0}}} [c(g,\pi_{g(x_t)})] \leq B$. We propose two methods to achieve this goal.

\subsection{Entropy-Based Policy Imitation (EPI)}
Our first method separates the learning of $g$ and $\pi_1$ as follows.
Fixing a low-cost architecture for pre-gating function $\tilde{g}$ and $\pi_1$, $\tilde{g}$ is trained to predict the entropy level of $\pi_0$ for any given state;  $\pi_1$ is trained to imitate the output distribution of $\pi_0$ for any given state. 
Specifically, we perform the following regressions:
\begin{gather}
\min_{\tilde{g} \in \mathcal{G}} \mathbb{E}_{x \in \mathcal{D}_{\pi_0}} ( \tilde{g}(x) - S(\pi_0 (x)) )^2, \\
\min_{\pi_1 \in \mathcal{F}} \mathbb{E}_{x \in \mathcal{D}_{\pi_0}} D(\pi_1(x)\| \pi_0 (x)),
\end{gather}
where $S(\cdot)$ denotes the entropy function and $D(\cdot \| \cdot)$ is the Kullback–Leibler (KL) divergence. Functions $\tilde{g}$ and $\pi_1$ can be either a logistic function, or a decision trees, or a deep neural network.

Next, we make the following observation. Given any state $x$, the entropy $S(\pi_0(x))$ of $\pi_0(x)$ can be quantized into 2 levels: high and low. Similarly, $S(\pi_1(x))$ can be quantized into high and low levels. When $S(\pi_0(x))$ is high, all actions are equally good; so any low-cost policy $\pi_1$ can be used instead without hurting long-term reward. When $S(\pi_0(x))$ is low, there is a clear winning action to take; furthermore, if $S(\pi_1(x))$ is low, we can expect it to imitate $\pi_0$ to pick out the correct action, in which case $x$ should be sent to $\pi_1$. If, however, $S(\pi_1(x))$ is high, we are less certain about the outcome of using $\pi_1$, and we prefer to send $x$ to $\pi_0$.
The above observation suggests two thresholding strategies based on the predicted entropy from $g(x)$ and $S(\pi_1(x))$. The first strategy \eqref{eq:EPI-1} is based on a single threshold on $\tilde{g}(x)$:
\begin{equation} \tag{EPI-1}\label{eq:EPI-1}
g(x) = \left\{
\begin{array}{rl}
0 & \text{if } \tilde{g}(x)\leq T_1 ,\\
1 & \text{if } \tilde{g}(x)>T_1,
\end{array} \right.
\end{equation}
where the hyperparameters $T_1$ can be chosen by simulation so that the budget constraint is satisfied. The above strategy simply sends examples to $\pi_1$ whenever $\pi_0$ is estimated to be uncertain. Our second entropy-based strategy \eqref{eq:EPI-2} has two thresholds:
\begin{equation} \tag{EPI-2}\label{eq:EPI-2}
g(x) = \left\{
\begin{array}{rl}
0 & \text{if } \tilde{g}(x)\leq T_1 \text{ and } S(\pi_1(x)) \leq T_2,\\
1 & \text{if } \tilde{g}(x)>T_1,
\end{array} \right.
\end{equation}
where the hyperparameters $T_1$ and $T_2$ can be chosen by simulation that the budget constraint is satisfied. 

EPI is based on the assumption that we know the good policy $\pi_0$ exactly, since entropy of $\pi_0$ is necessary. The good policy may be obtained directly by policy gradient method and A3C algorithms mentioned above. For DQN or double-DQN, the technique in \cite{parisotto2015actor} can be applied to transform the DQN into a policy network by
\begin{equation*}
\pi (a|x) = \dfrac{e^{Q(x, a)}}{\sum_{a' \in \mathcal{A}} Q(x, a')},
\end{equation*}
so that EPI can be applied.

\subsection{Alternating Minimization Policy imitation (API)}
The entropy-based policy imitation algorithms are easy to implement and straight forward. Yet, finding the correct hyperparameters may be expensive in some situations. Therefore, we propose an adaptive approximation algorithm based on \cite{nan2017dynamic}. Although their formulation applies to the batch setting, we can adapt it to our setting via sampling mini-batches based on $\pi_0$. As before, we fix the good policy $\pi_0$ as input. There are three components to learn: $\pi_1$ is the weak policy; $q(\cdot|x) \in [0,1]$ is an intermediate gating likelihood function; $g$ is the gating function, which assumes a logistic model i.e., $g(x)=\sigma(\tilde{g}) = 1/(1+\exp(-\tilde{g}))$.
The overall optimization problem to solve is
\begin{align}
\label{eq:em_all}
\begin{split}
\min_{\mu_{1} \in \mathcal{F}, g \in \mathcal{G}, q} & \mathbb{E}_{x \in \mathcal{D}_{\pi_0}} q(1|x) D(\pi_1(x)\| \pi_0(x)) + D(q(\cdot| x)\| g(x)) + \Omega(\pi_1, g) \\
\text{subject to}: \quad& \mathbb{E}_{x \in \mathcal{D}_{\pi_0}} [q(0|x)]\leq P_{\rm full}
\end{split}
\end{align}
where $x \in \mathcal{D}_{\pi_0}$ comes from the demonstration trajectory of the good policy, $D(\cdot \| \cdot)$ is again KL divergence and $\Omega$ is a penalty (e.g., an $L_2$ type of penalty form) on the complexity of $g$ and $\pi_1$. The constraint ensures that only a $P_{\rm full}$ fraction of state samples are sent to the good policy $\pi_0$. 

We use alternating minimization to solve the above problem. First, we fix a value for $\pi_1$ and $g$, and solve for $q$ in \eqref{eq:em_all}. In particular, for a batch of demonstrations $\{x_i\}_{i=1}^N$, let $q_i = q(0|x_i)$. It turns out that there is a closed-form solution:  $q_i = 1/(1 + e^{B_i - A_i + \beta})$, where $A_i = \sigma^+(\tilde{g}(x_i)) + D(\pi_1(x_i),\| \pi_0(x_i))$, $B_i = \sigma^+(-\tilde{g}(x_i))$, $\sigma^+(x) = \log(1 + e^x)$, and $\beta$ is a non-positive number such that the constraint is satisfied.

Next, we fix the $q$'s and optimize for $\pi_1$ and $g$:
\begin{align}
\min_{\pi_{1} \in \mathcal{F}, g \in \mathcal{G}} \sum_{i=1}^N (1-q_i) D(\pi_1(x)\| \pi_0(x)) + D(q_i\| g(x)) + \Omega(\pi_1, g). \label{eq:m_step}
\end{align}
We state the detailed algorithm as follows.
\begin{algorithm}[H]
	\caption{Alt-Min Policy Imitation (API) for MDP}
	\label{alg:train}
	\begin{algorithmic}
		\STATE Initialize weak policy function $\pi_1$
		\STATE Initialize gating function $g$
		\FOR{epoch $\in \{1, 2, \cdots \}$} 
		\STATE reset batch memory $\mathcal{M}$
		\FOR{step $\in \{1, 2, \cdots N\}$} 
		\STATE Get state $x$ in the MDP.
		\STATE Perform action $a$ according to the good policy $\pi_0$.
		\STATE Store (x, $\pi_0(x)$) in batch memory $\mathcal{M}$
		\IF{MDP terminated}
		\STATE Restart MDP
		\ENDIF
		\ENDFOR
		\STATE Update $q_i = 1/(1 + e^{B_i - A_i + \beta})$ for $x_i$ in batch $\mathcal{M}$
		\STATE Perform a gradient decent step of loss \eqref{eq:m_step} under $\mathcal{M}$ and $q$
		\ENDFOR
	\end{algorithmic}
\end{algorithm}

\section{Experiments}
We evaluate the performance of our proposed algorithms in the Atari 2600 games in OpenAI Gym Environment \cite{gym}. For our experiment, we use a subset of seven games. We first obtained the good polices by using the A3C algorithm in \cite{mnih2016asynchronous}. To demonstrate the effectiveness of our proposed model, we set the following two baseline policies. We first train a weak policy using the A3C algorithm and use the following switching rule. The random switching rule assigns a $P_{\rm full}$ fraction of input states to the good policy and the rest to the weak policy. The naive switching rule assigns to the good policy the states for which the weak policy has a high entropy. The performances of the good and weak policies can be found in Table \ref{tab:atari}. Note the weak policy and the gating function shares the same convolutional layer and it is 6x faster than the good policy.

We notice that the game scores achieved by a good policy usually follow a heavy tail distribution, i.e., the score achieved when the game terminates usually has a very high variance. Also, we notice that most of the games repeat the same content after one round. We use the approximate game score achieved in the first round to evaluate the performance\footnote{We observe the average time that the good policy finishes one round for all of these games is about 2 minutes. Pong does not have rounds, but it usually ends before 2 minutes. So the policies are evaluated for 2 mins of emulator time (3, 000 frame).}. 
\begin{figure}[htbp]
	\centering
	\begin{subfigure}[b]{0.75\linewidth}
		\includegraphics[width=\textwidth]{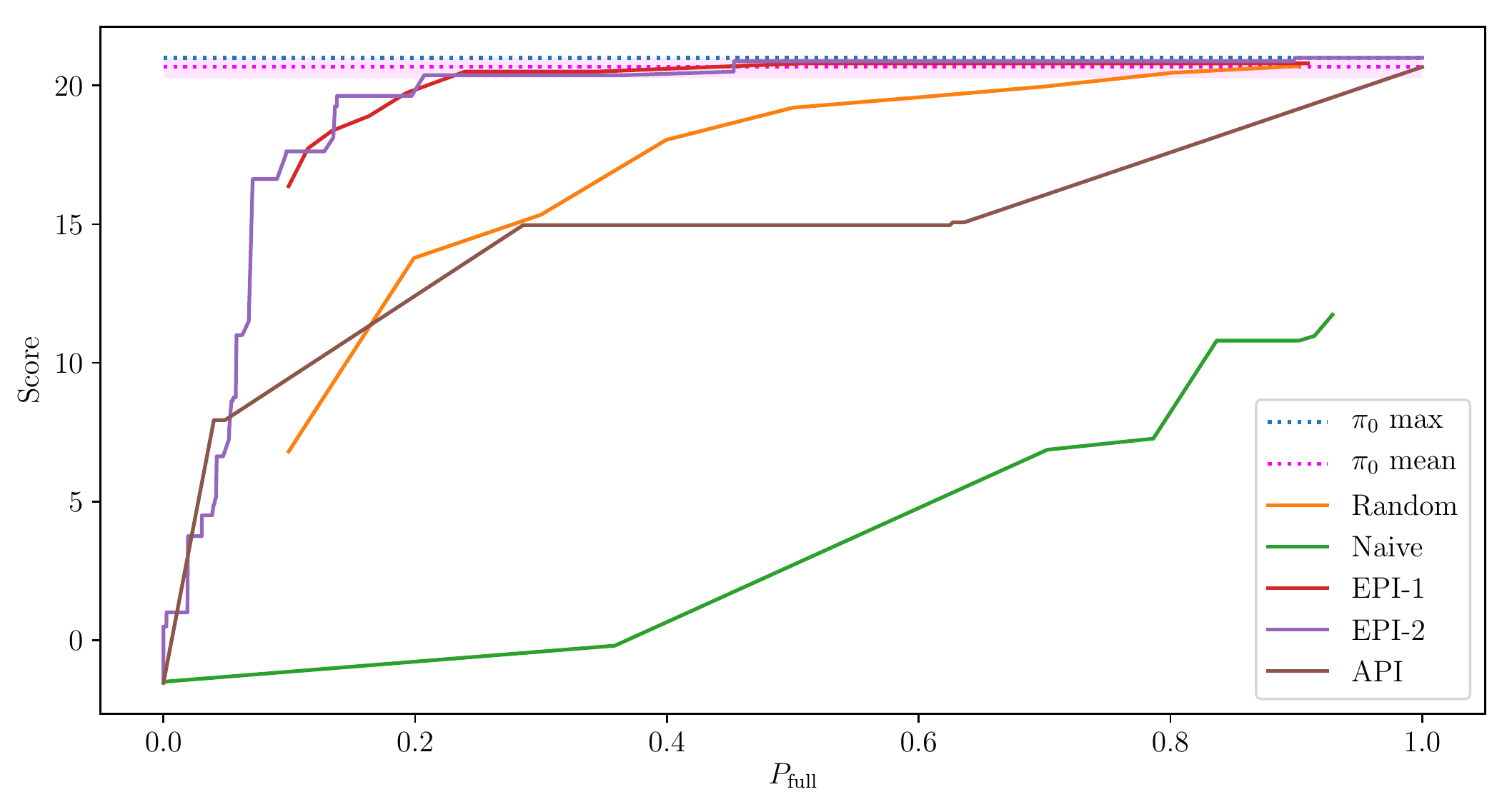}
		\caption{Game scores of Pong by different policies.}
		\label{fig:pong}
	\end{subfigure}
	
	\begin{subfigure}[b]{0.75\linewidth}
		\includegraphics[width=\textwidth]{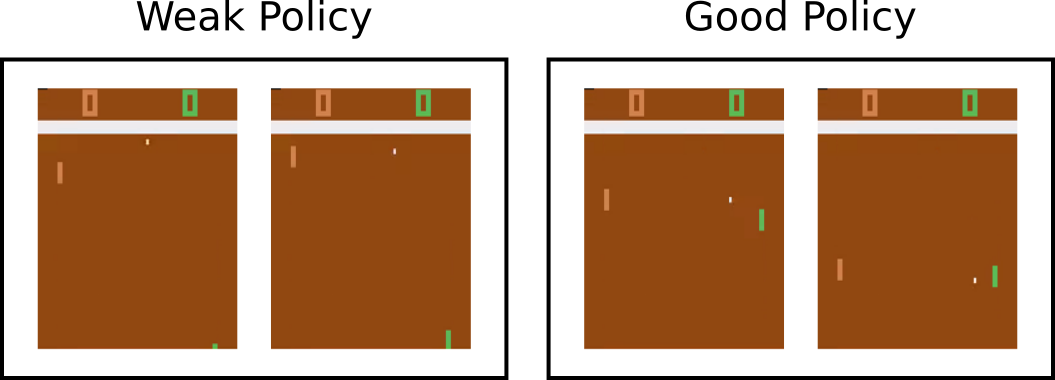}
		\caption{Sample gameplay by our entropy-based policy imitations.}
		\label{fig:gameplay}
	\end{subfigure}
    
    \begin{subfigure}[c]{0.75\linewidth}
        \includegraphics[width=\textwidth]{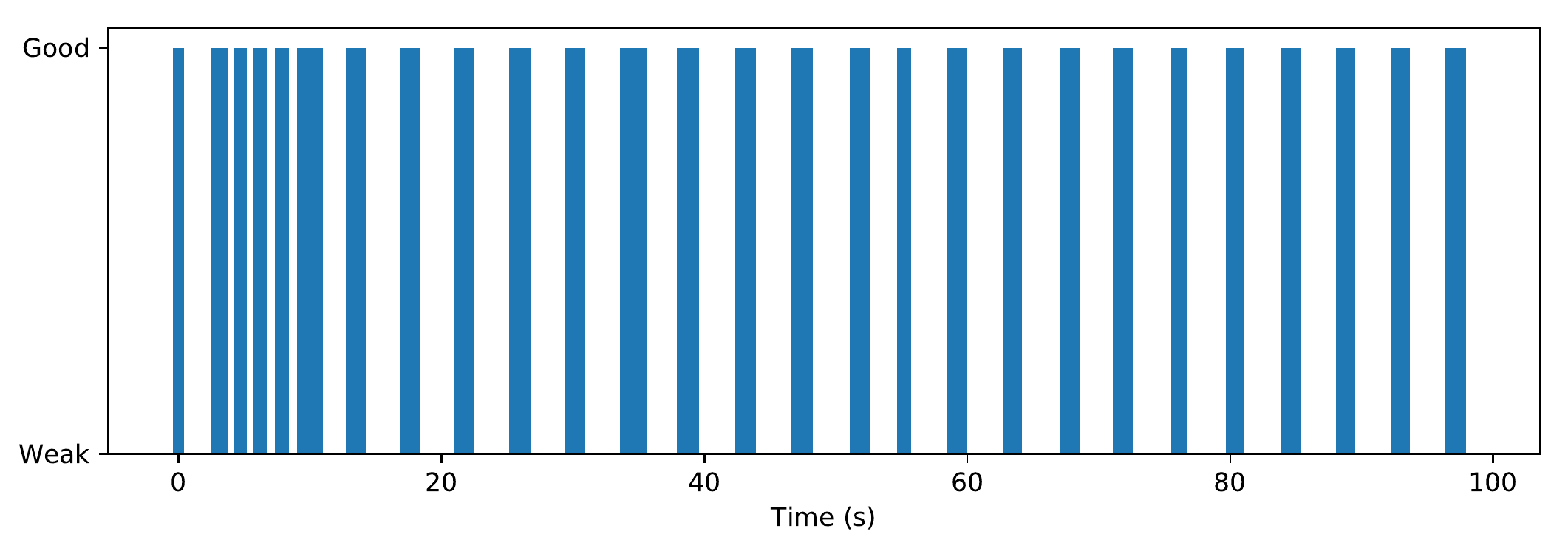}
		\caption{Selection of the two policies during the gameplay.}
		\label{fig:test_time}
	\end{subfigure}
    
	\caption{\textbf{(a)} Comparison of different policies on Pong under different $P_{\rm full}$. The curves shows averages over 20 evaluation episodes at 2 min. Our entropy-based policy imitation methods significantly outperform the rest of the methods. By assigning less than $20\%$ states to the good policy, the EPI-2 achieves a similar performance to the good policy. The magenta region is the score between the maximum score achieved by the good policy and one standard deviation below the mean.
		\textbf{(b)} The sample gameplay images generated by our EPI-2. The gating function selects the weak policy for the states where the ball is far away from the paddle controlled by the agent on the right side of the screen. At the time when the ball is close to the paddle, the gating function selects the good policy.
        \textbf{(c)} Selection of the good and weak policies during the sample gameplay. An illustration of the gameplay using EPI-2 can be seen in the supplementary video.}
\end{figure}

\begin{figure}[htbp]
	\centering
	\begin{subfigure}[b]{0.45\linewidth}
		\includegraphics[width=\textwidth]{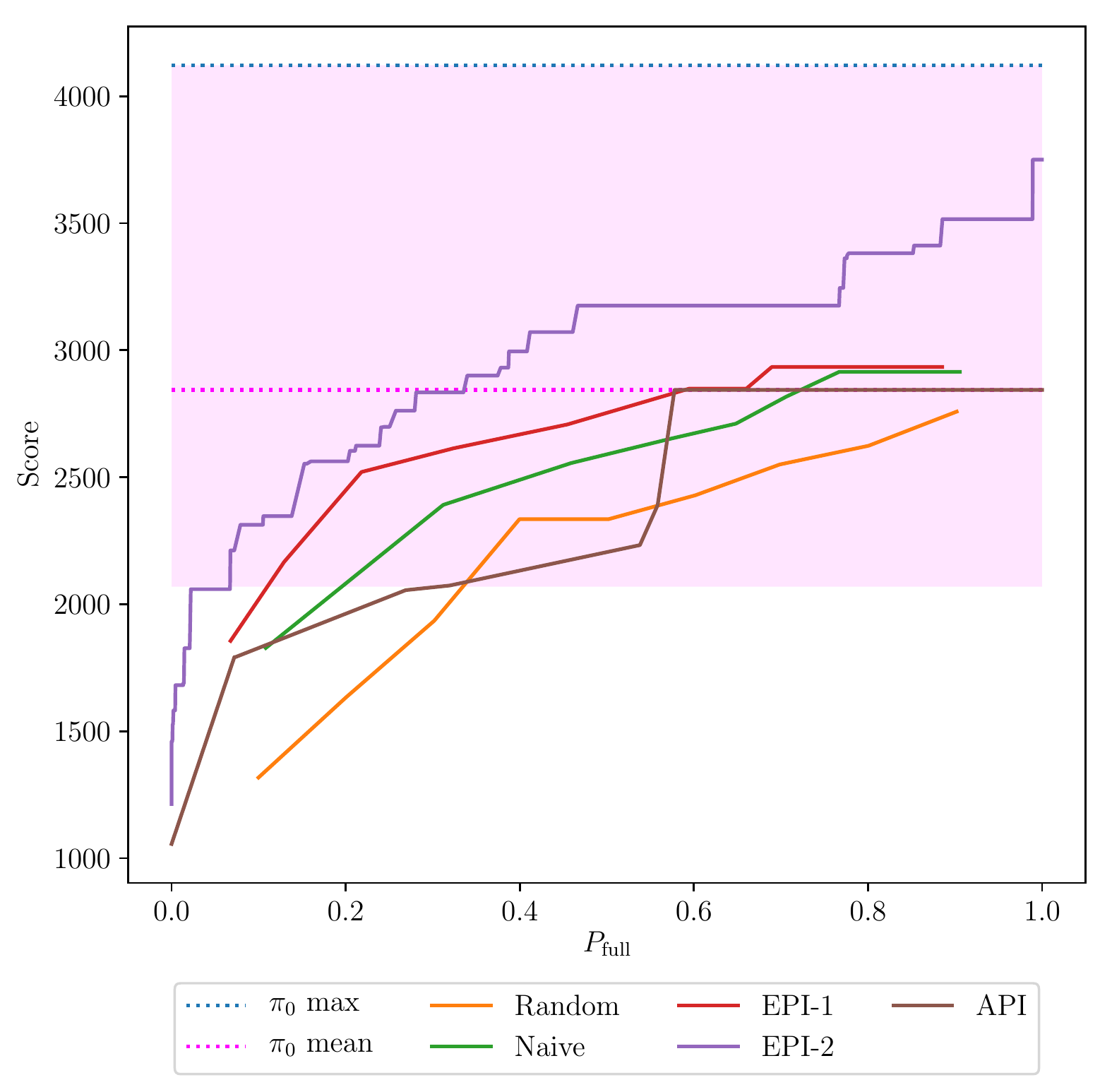}
		\caption{Assault}
	\end{subfigure}
	\hfill
	\begin{subfigure}[b]{0.45\linewidth}
		\includegraphics[width=\textwidth]{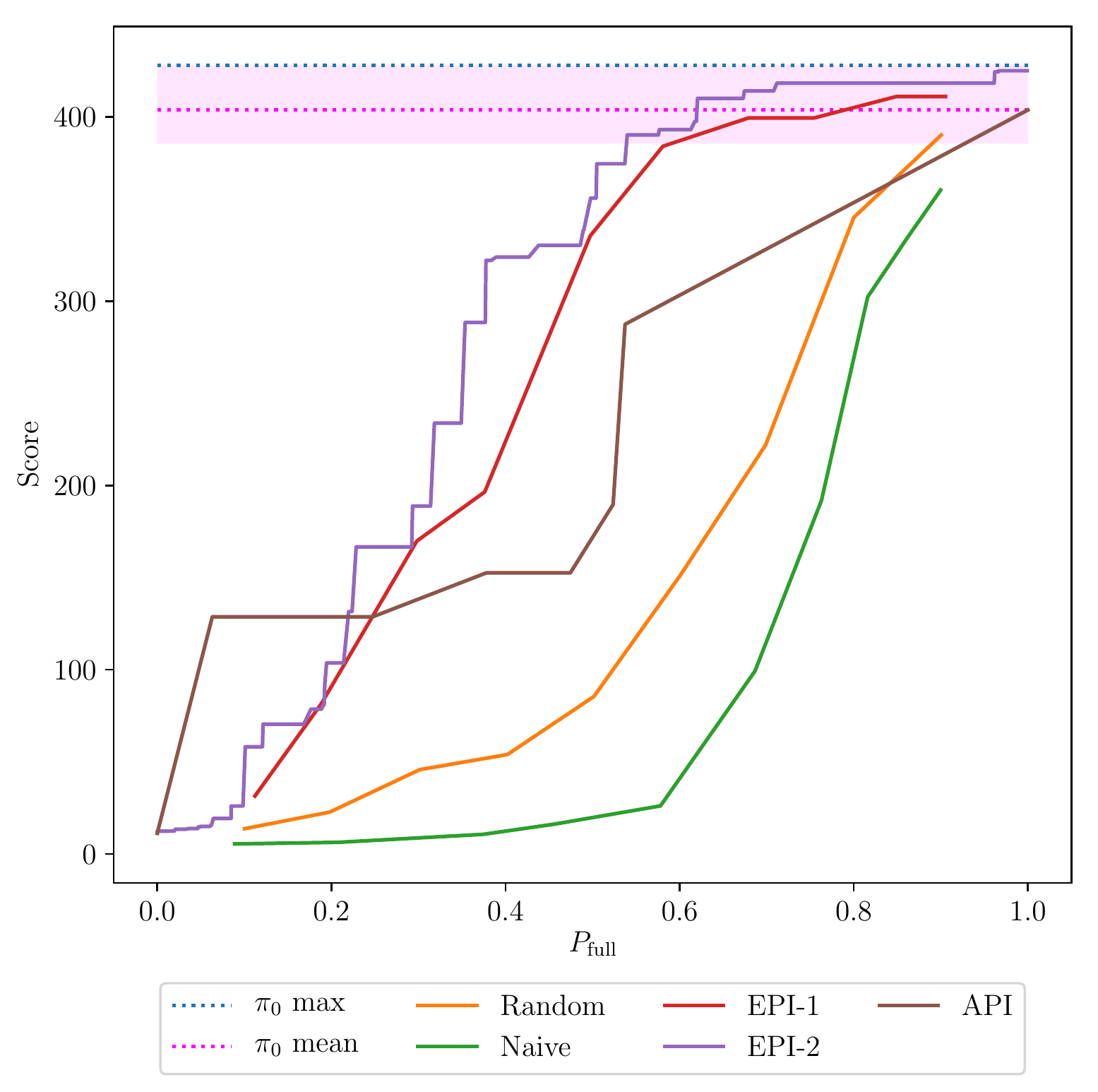}
		\caption{Breakout}
	\end{subfigure}
	
	\begin{subfigure}[b]{0.45\linewidth}
		\includegraphics[width=\textwidth]{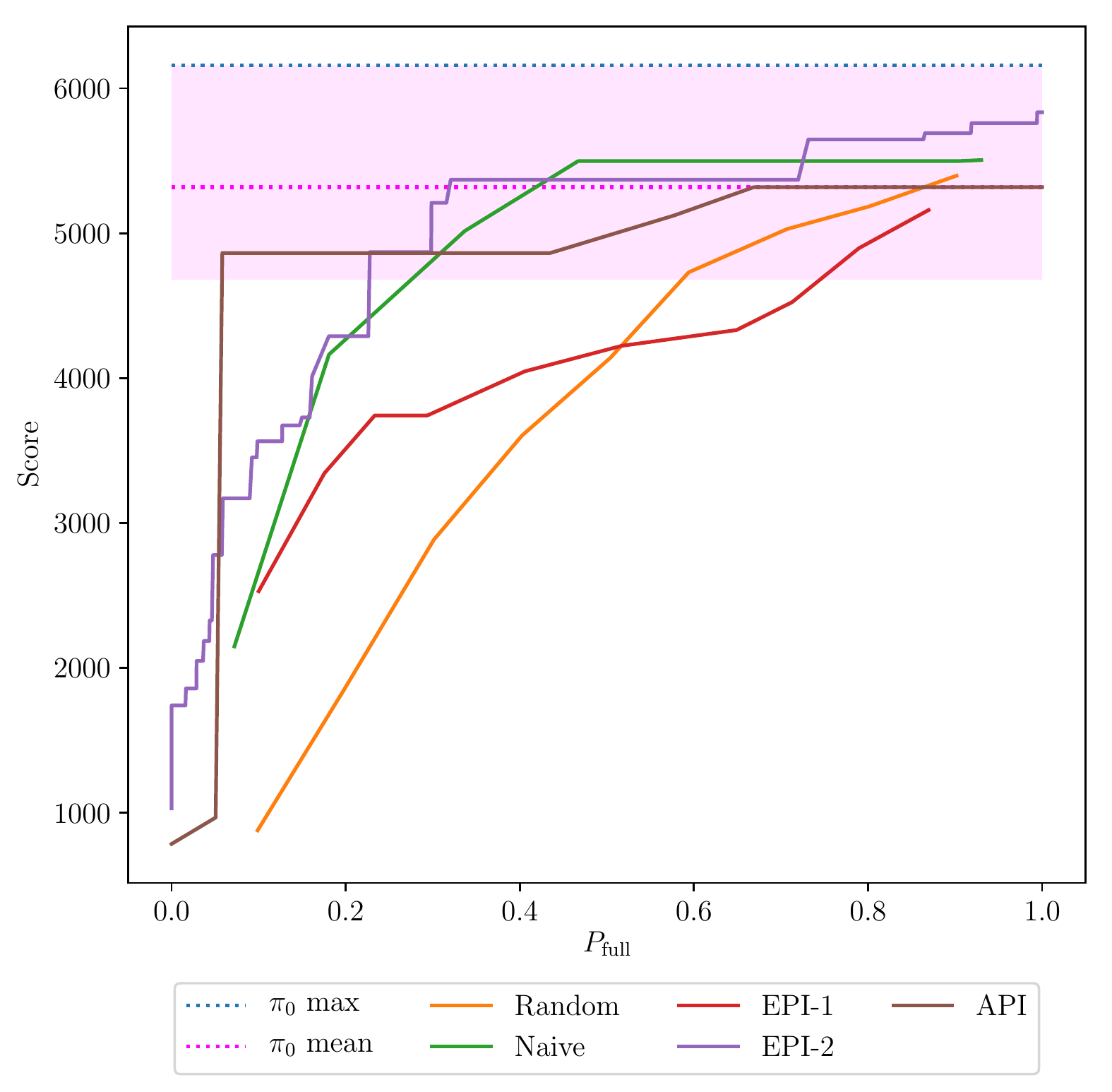}
		\caption{Carnival}
	\end{subfigure}
	\hfill
	\begin{subfigure}[b]{0.45\linewidth}
		\includegraphics[width=\textwidth]{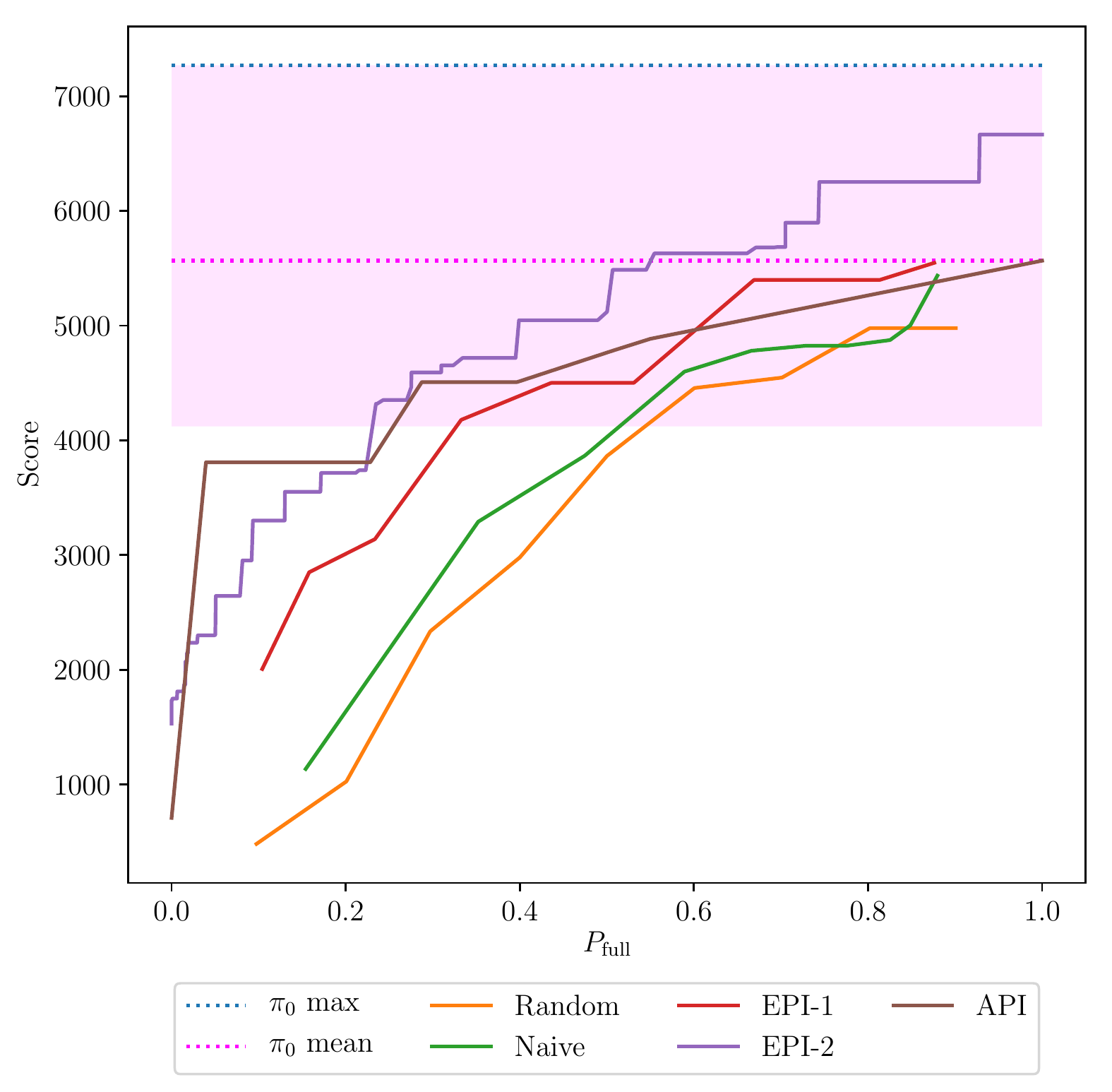}
		\caption{DemonAttack}
	\end{subfigure}
	
	\begin{subfigure}[b]{0.45\linewidth}
		\includegraphics[width=\textwidth]{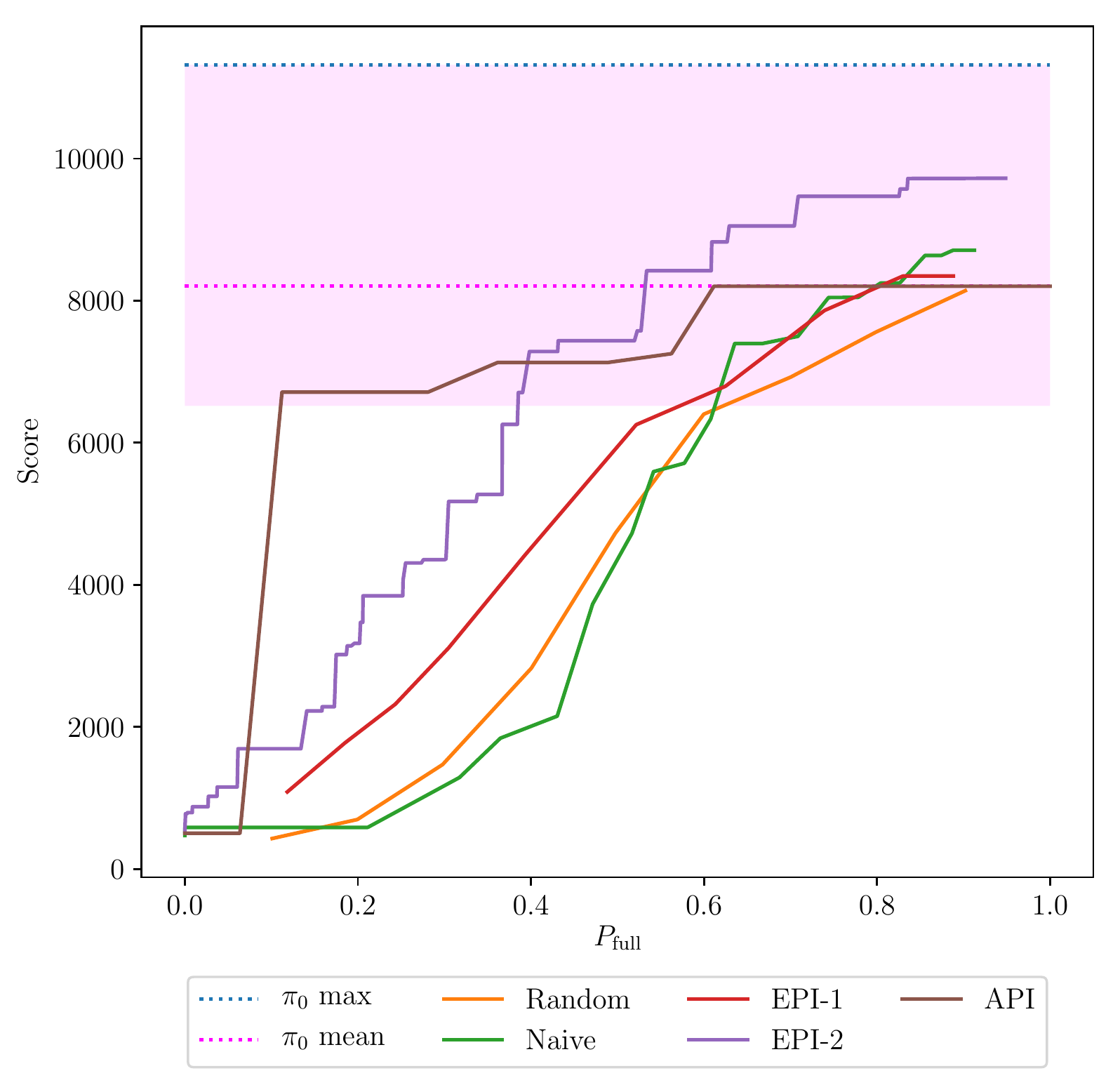}
		\caption{Seaquest}
	\end{subfigure}
	\hfill
	\begin{subfigure}[b]{0.45\linewidth}
		\includegraphics[width=\textwidth]{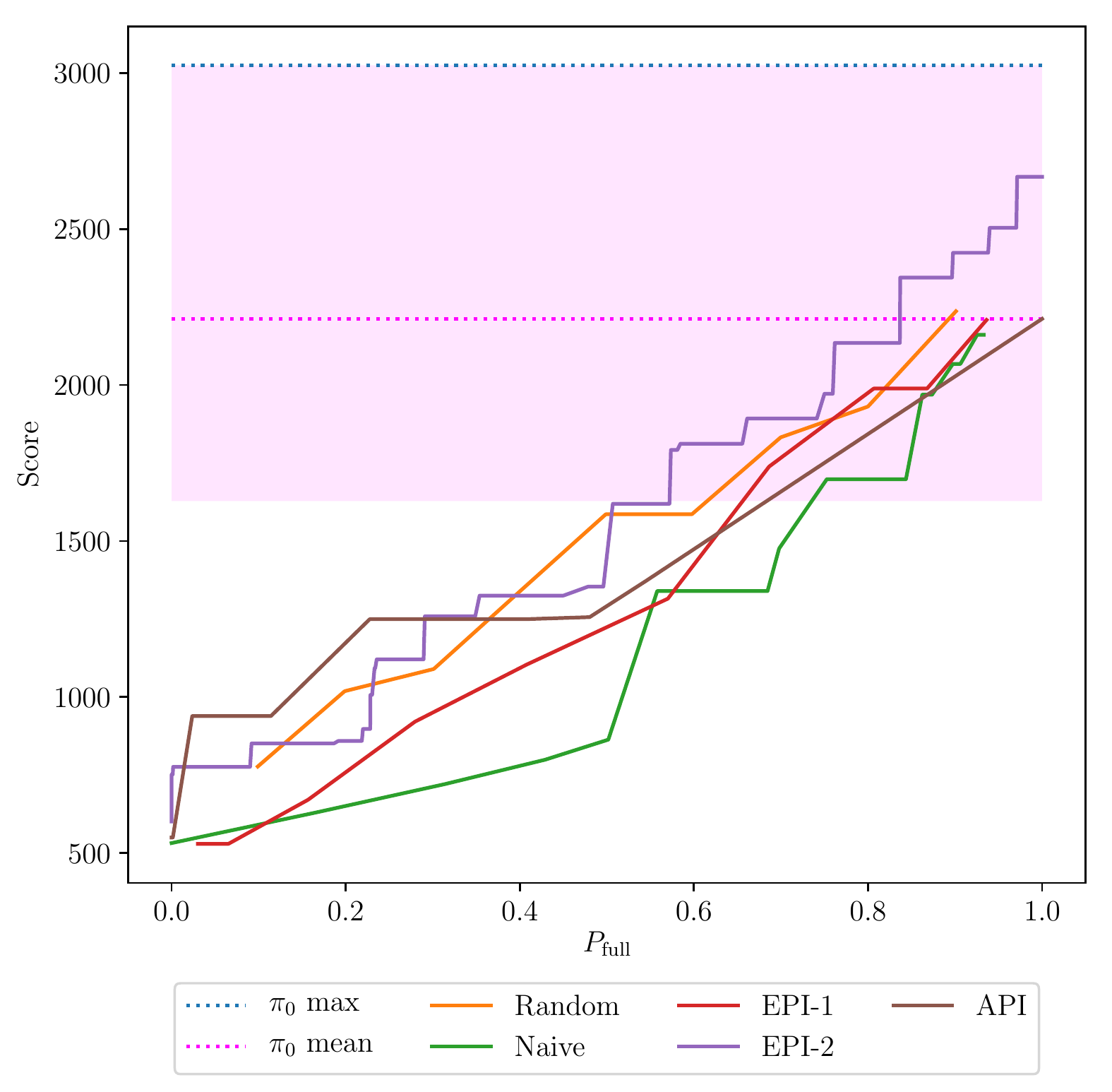}
		\caption{Space Invaders}
	\end{subfigure}
	\caption{The performance of different kinds of policies for different games. The magenta region is the score between the maximums score achieved by the good policy and one standard deviation below the mean. The curves show averages over the 20 runs. For API method, we only set $P_{\rm full}$ to be $0.1$, $0.3$, and $0.5$, and train with 3 different $L_2$ regularization parameters: $0$, $10^{-5}$ and $10^{-3}$. }
	\label{fig:result}
\end{figure}

The performance of the imitation polices introduced in this paper and the baseline ones are shown in Figures \ref{fig:pong} and \ref{fig:result}. For five out of the seven games, we can achieve similar level of performance by assigning less than $30\%$ of the state to the good policy. This implies that we achieve a more than 2x speed-up during test-time. In the most extreme case, Carnival, API only assigns $0.5\%$ states to the good policy and maintains similar performance as the good one. This represents an almost 5x speed-up compared to the good policy. However, for the game Breakout and Space Invaders, we need to assign no more than half of the states to the weak policy to achieve a similar performance. 

EPI-2 usually has a very stable performance under different level of $P_{\rm full}$. This policy always outperforms the baseline methods for different $P_{\rm full}$. Pong is an interesting example that helps explain how the gating function in EPI-2 works. EPI-2 can assign less than $20\%$ of input states to the good policy while achieving a similar performance as the good one. Figure \ref{fig:gameplay} shows the gameplay of the Pong under the EPI-2 policy. We find that the gating function selects the weak policy for the states where the ball is far away from the  paddle controlled by the agent on the right side of the screen. The weak policy does not even align the paddle with the ball horizontally. At the time when the ball is close to the paddle, the gating function selects the good policy. Note the input of the gating function is the current image. So even when the ball begins to move away from the paddle, it will still choose the good policy. This observation indicates the gating function is learned to select the state that is critical to obtain a higher score. Since the gating function is based on the entropy of the good policy, this result is consistent with the observation in Extended Data Figure 2 in \cite{mnih2015human}. 

\begin{table}[htbp]
	\begin{center}
		\begin{tabular}{|c|c|c|c|c|c|c|c|c|}
			\hline
			$P_{\rm full}$ &  & Assault & Breakout & Carnival & Dem. Att. & Pong & Seaquest & Sp. Inva.\\
			\hline
			& Good & 2,741.33 & 403.70 & 5,365.00 & 5,726.00 & 21.00 & 8,164.33 & 2,333.50\\ \hline
			& Weak & 1,056.30 & 11.40 & 708.00 & 156.50 & 1.40 & 504.00 & 531.50\\ \hline
			\multirow{3}{*}{0.1} & Random & 1,317.52 & 20.90 & 648.00 & 482.10 & 7.50 & 408.00 & 777.20\\ \cline{2-9}
			& \multirow{ 2}{*}{API} & \textbf{1,791.93} & \textbf{128.63} & \textbf{4,973.00} & \textbf{3,808.83} & \textbf{7.93} & \textbf{6,711.33} & \textbf{938.83}\\
			&  & (0.07) & (0.06) & (0.05) & (0.04) & (0.05) & (0.11) & (0.02)\\ \hline
			\multirow{3}{*}{0.3} & Random & 1,934.98 & 35.60 & 2,768.00 & 2,335.30 & \textbf{15.40} & 1,412.00 & 1,089.70\\ \cline{2-9}
			& \multirow{ 2}{*}{API} & \textbf{2,073.50} & \textbf{200.00} & \textbf{5,010.00} & \textbf{4,507.67} & 14.97 & \textbf{7,128.67} & \textbf{1,249.50}\\
			&  & (0.32) & (0.34) & (0.34) & (0.29) & (0.28) & (0.36) & (0.25)\\ \hline
			\multirow{3}{*}{0.5} & Random & 2,165.30 & 61.80 & 3,837.00 & 3,534.00 & \textbf{18.60} & 4,465.00 & \textbf{1,585.80}\\ \cline{2-9}
			& \multirow{ 2}{*}{API} & \textbf{2,843.47} & \textbf{287.40} & \textbf{5,474.00} & \textbf{4,884.83} & 14.80 & \textbf{8,202.33} & 1,372.00\\
			&  & (0.58) & (0.54) & (0.60) & (0.55) & (0.64) & (0.62) & (0.54)\\ \hline
		\end{tabular}
	\end{center}
	\caption{Average game scores of the different policies. The scores are the averages over 20 runs. Using the specified $P_{\rm full}$'s in API algorithm, the proportion of the states that is sent to the good policy in the test case, are shown in the parentheses below the score. For a fix $P_{\rm full}$, we train API with three different $L_2$ regularization parameters: $0$, $10^{-5}$ and $10^{-3}$. The weak policy $\pi_1$ is the best policy learned with the weak architecture stated in the supplementary document. For each game, we use bold font for the policy result with the higher average game score for particular column.}
	\label{tab:atari}
\end{table}

The second method we proposed, API,  has a good performance when $P_{\rm full}$ is small, but is not stable when $P_{\rm full}$ is large. For six out of seven games, at $P_{\rm full}=0.1$ level, API outperforms every other method. We believe when the $P_{\rm full}$ is small, the entropy is not a good indicator of the importance of the state. We also list the result of the API under 3 levels of $P_{\rm full}$, 0.1, 0.3, and 0.5 respectively. The proportion of states that is sent to the good policy in the test cases is similar to the $P_{\rm full}$ specified in the algorithm.

One interesting observation is that the EPI methods achieve a higher score than the good policy under a large $P_{\rm full}$. Admittedly, this may due to the randomness of the game evaluation, since the good policies still have a high variance. We, however, would suggest that the good policy may over generalize in the test period, since its model complexity is high. Replacing the good policy with high uncertainty with a weak policy, which has a low model complexity and is less likely to overfit, serves as a regularization process for the good policy.

\section{Related Work}
Resource-constrained machine learning has become an active area of research. Many algorithms have been proposed to reduce test-time costs (in terms of computation or feature acquisition) for classification/ regression \cite{Gao+Koller:NIPS11,wang2014lp,trapeznikov:2013b,DBLP:conf/icml/XuWC12, icml2015_nan15,NanNIPS2016,NIPS2015_5982} and structured prediction \cite{DBLP:conf/aaai/BolukbasiCWS17} by learning adaptive feature acquisition/evaluation rules using generative models or empirical risk minimization with no direct relation to RL.

Others have formulated the adaptive feature acquisition/evaluation problem as a MDP \cite{Dulac-Arnold:2011:DCS:2034063.2034100,DBLP:conf/icml/Busa-FeketeBK12,Karayev_dynamicfeature,he2012imitation}. 
They encode observations so far as state, unused features/base classifiers as action space, and formulate various reward functions to account for classification error and costs. Such methods have been successfully applied in NLP \cite{he2013dynamic} to adaptively select features for dependency parsing, as well as computer vision \cite{weiss2013learning} for human pose tracking. These methods use RL as a tool to learn decision rules that reduce test-time cost in the static environment of classification or structured prediction problems. 

In contrast, our objective is to improve the latency of RL itself in a dynamic environment rather than utilizing RL as a tool. We need to make sequential decisions based on dynamically evolving states. Our goal is to dynamically switch between low complexity shallow NNs and high-complexity deep NNs so that latency is reduced but we suffer no loss in terminal value. This notion of terminal value and seeking actions for maximal future reward and doing so in cost-effective manner is our novel contribution.

The algorithms proposed in this paper use techniques from imitation learning or Learning from Demonstration (LfD), where a expert's demonstration or guidance is used to help learn the MDP. One of such methods is DAGGER \cite{ross2011reduction}, which trains a policy that directly imitate the expert's behavior. Recent works have exploited this idea by training a multi-task reinforcement learning agent \cite{parisotto2015actor} or perform deep Q-learning from demonstrations \cite{DQfD}. These can be viewed as complementary to our work as we use these tools to reduce latency in policy evaluation. 
 
The teacher-student or distilling framework \cite{LopSchBotVap16,romero2014fitnets, hinton2015distilling} is also related to our approach; a low-cost student policy model learns to approximate the teacher policy model so as to meet test-time budget. However, the goal there is to learn a better stand-alone student model. In contrast, we make use of both the low-cost (student) and high-accuracy (teacher) policy models during prediction via a gating function, which learns the limitation of the low-cost (student) model and consults the high-accuracy (teacher) model if necessary, thereby avoiding accuracy loss. 

Finally, our approach draws inspiration from cognitive science and neuroscience. \cite{daw2005uncertainty} suggests that there are two systems in the brain, being coordinated by the uncertainty of the decision under a competition rule. \cite{kahneman2011thinking} implies that the switching of two systems that are with different levels of accuracy, reflect time, and metabolic cost, saves human energy budgets while maintaining a moderately good decision-making ability. This is reminiscent of our combination of gating, SNN and DNN to meet budget constraints. 

\section{Conclusions}
In this paper, we proposed two imitation learning algorithms. They mimic a good policy by adaptively select good or weak policies so that the adaptive switching policy has a similar performance with the good policy. One of our proposed methods uses a gating function to estimates the entropy of the good policy, and the second one formulates a joint margin-based optimization objective and solves it with an alternating minimization scheme. We demonstrate a significant reduction in the test-time for the Atari games. A direction of future work is to develop stable methods that can learn a gating function and a weak policy through reinforcement learning methods. Nevertheless, using imitation learning can help in cases of obtaining a fast test-time policy for complicated tasks.

\bibliographystyle{apalike}
\bibliography{sample}
\clearpage
\section{Appendix: Experimental Details}
\begin{figure}[htbp]
	\centering
	\begin{subfigure}[b]{0.40\linewidth}
		\includegraphics[width=\textwidth]{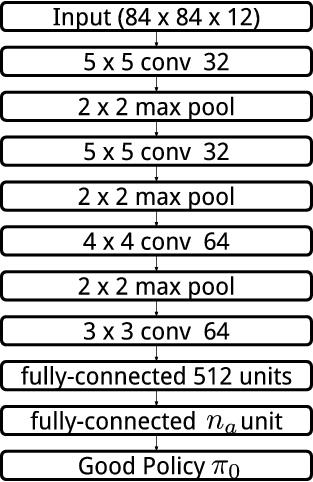}
		\caption{Good Policy.}
		\label{fig:good_net}
	\end{subfigure}
	\hfill
	\begin{subfigure}[b]{0.55\linewidth}
		\includegraphics[width=\textwidth]{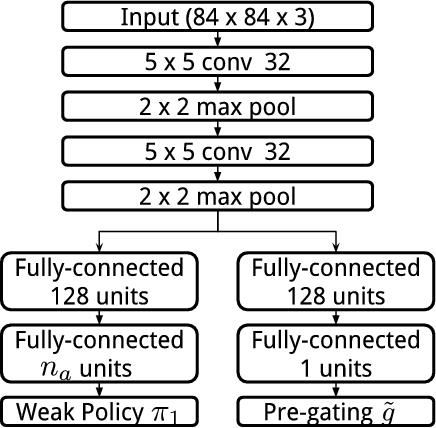}
		\caption{Weak policy $\pi_1$ and pre-gating function $\tilde g$.}
		\label{fig:weak_net}
	\end{subfigure}
	\caption{The neural network structure of the good policy $\pi_0$, weak policy $\pi_1$ and pre-gating function $\tilde g$ for Atari game. Here, $n_a$ is the number of the action in the game.}
\end{figure}

We use the OpenAI Gym environment \cite{gym} to simulate the Atari games. Specifically, we choose Assault, Breakout, Carnival, Demon Attack, Pong, Seaquest, and Space Invaders among the Atari game environments. We resize the image of the game into $84 \times 84$ pixels. The good policy is a convolutional neural network with the most recent 4 images as the input. So the input tensor has shape $84 \times 84 \times 12$. The first hidden layer convolves $32$ filters of $5 \times 5$ of stride $1$ and applies an ReLU function. The second layer is a max pooling layer of $2 \times 2$ with stride 2. The third layer and the fourth layer are the same as the first layer and the second layer. The fifth layer convolves $64$ filters of $4 \times 4$ of stride $1$ and applies an ReLU function. The six layer is a max pooling layer of $2 \times 2$ with stride 2. This follows a fourth convolutional layer convolving $64$ filters $3 \times 3$ with ReLU activation function. The final hidden layer id fully-connected and consists of 512 rectifier units. The output layer is a fully-connected layer with $n_a$ softmax units, where $n_a$ is the number of action. Figure \ref{fig:good_net} shows the diagram of the good policy. The weak policy, entropy estimate and pre-gating function are also convolutional neural networks. They share the same convolutional and max pooling layers. The input for the neural network is the most recent image. So the input tensor is of $84 \times 84 \times 3$. The first hidden layer convolves $16$ filters of $5 \times 5$ of stride $1$ and applies an ReLU function. The second layer is a max pooling layer of $2 \times 2$ with stride 2. The third layer and the fourth layer are the same as the first layer and the second layer. This ends the sharing part of the neural networks. For the weak policy, it follows a last hidden layers with 128 rectifier units. The output layer is a fully-connected layer with $n_a$ softmax units. Figure \ref{fig:good_net} shows the diagram of the weak policy and the pre-gating function. For the entropy estimate or the pre-gating function, it also follows the last hidden layers with 128 rectifier units. The output layer is a fully-connected layer with a single neuron. For the good policy $\pi_0$, it takes about 132M multiplication operations for a single image input, while for the weak policy $\pi_1$ takes only less than 20M.

We use the A3C (Algorithm S3 in \cite{mnih2016asynchronous}) to obtain the good and weak policies for every selected game. For our algorithm, we uses the Adam optimization algorithm \cite{kingma2014adam} with learning rate $lr=0.0005$, $\beta_1=0.9$, $\beta_2=0.999$, and $\epsilon=10^{-8}$. The batch-size in the training is $1000$ and each game takes $10, 000$ iterations. For evaluation of the game score, we repeat the game 20 times to find the average score of the game. For API method, we only set $P_{\rm full}$ to be $0.1$, $0.3$, and $0.5$, and train with 3 different $L_2$ regularization parameters: $0$, $10^{-5}$ and $10^{-3}$. The EPI is also trained with 3 different $L_2$ regularization parameters: $0$, $10^{-5}$ and $10^{-3}$.

\end{document}